\documentclass{article}

\usepackage[numbers]{natbib}
\usepackage[final]{nips_2017} 

\usepackage[utf8]{inputenc} 
\usepackage[T1]{fontenc}    
\usepackage{hyperref}       
\usepackage{url}            
\usepackage{booktabs}       
\usepackage{amsfonts}       
\usepackage{nicefrac}       
\usepackage{microtype}      
\usepackage{graphicx}
\usepackage{float}
\usepackage{subcaption}
\usepackage{caption}
\usepackage{multirow}
\usepackage[fleqn]{amsmath}

\title{pix2code: Generating Code from a Graphical User Interface Screenshot}

\author{
    Tony Beltramelli \\
    UIzard Technologies\\
    Copenhagen, Denmark\\
    \texttt{tony@uizard.io}
}

\begin{document}

\maketitle

\begin{abstract}
Transforming a graphical user interface screenshot created by a designer into computer code is a typical task conducted by a developer in order to build customized software, websites, and mobile applications. In this paper, we show that deep learning methods can be leveraged to train a model end-to-end to automatically generate code from a single input image with over 77\% of accuracy for three different platforms (i.e. iOS, Android and web-based technologies).
\end{abstract}

\section{Introduction}

The process of implementing client-side software based on a \emph{Graphical User Interface (GUI)} mockup created by a designer is the responsibility of developers. Implementing GUI code is, however, time-consuming and prevent developers from dedicating the majority of their time implementing the actual functionality and logic of the software they are building. Moreover, the computer languages used to implement such GUIs are specific to each target runtime system; thus resulting in tedious and repetitive work when the software being built is expected to run on multiple platforms using native technologies. In this paper, we describe a model trained end-to-end with stochastic gradient descent to simultaneously learns to model sequences and spatio-temporal visual features to generate variable-length strings of tokens from a single GUI image as input.

Our first contribution is \emph{pix2code}, a novel approach based on Convolutional and Recurrent Neural Networks allowing the generation of computer tokens from a single GUI screenshot as input. That is, no engineered feature extraction pipeline nor expert heuristics was designed to process the input data; our model learns from the pixel values of the input image alone. Our experiments demonstrate the effectiveness of our method for generating computer code for various platforms (i.e. iOS and Android native mobile interfaces, and multi-platform web-based HTML/CSS interfaces) without the need for any change or specific tuning to the model. In fact, \emph{pix2code} can be used as such to support different target languages simply by being trained on a different dataset. A video demonstrating our system is available online\footnote{\url{https://uizard.io/research\#pix2code}}.

Our second contribution is the release of our synthesized datasets consisting of both GUI screenshots and associated source code for three different platforms. Our datasets and our \emph{pix2code} implemention are publicly available\footnote{\url{https://github.com/tonybeltramelli/pix2code}} to foster future research.

\section{Related Work}

The automatic generation of programs using machine learning techniques is a relatively new field of research and program synthesis in a human-readable format have only been addressed very recently. A recent example is DeepCoder \cite{balog2016deepcoder}, a system able to generate computer programs by leveraging statistical predictions to augment traditional search techniques. In another work by Gaunt et al. \cite{gaunt2016terpret}, the generation of source code is enabled by learning the relationships between input-output examples via differentiable interpreters. Furthermore, Ling et al. \cite{ling2016latent} recently demonstrated program synthesis from a mixed natural language and structured program specification as input. It is important to note that most of these methods rely on \emph{Domain Specific Languages (DSLs)}; computer languages (e.g. markup languages, programming languages, modeling languages) that are designed for a specialized domain but are typically more restrictive than full-featured computer languages. Using DSLs thus limit the complexity of the programming language that needs to be modeled and reduce the size of the search space.

Although the generation of computer programs is an active research field as suggested by these breakthroughs, program generation from visual inputs is still a nearly unexplored research area. The closest related work is a method developed by Nguyen et al. \cite{nguyen2015reverse} to reverse-engineer Android user interfaces from screenshots. However, their method relies entirely on engineered heuristics requiring expert knowledge of the domain to be implemented successfully. Our paper is, to the best of our knowledge, the first work attempting to address the problem of user interface code generation from visual inputs by leveraging machine learning to learn latent variables instead of engineering complex heuristics.

In order to exploit the graphical nature of our input, we can borrow methods from the computer vision literature. In fact, an important number of research \cite{vinyals2015show, donahue2015long, karpathy2015deep, xu2015show} have addressed the problem of image captioning with impressive results; showing that deep neural networks are able to learn latent variables describing objects in an image and their relationships with corresponding variable-length textual descriptions. All these methods rely on two main components. First, a \emph{Convolutional Neural Network (CNN)} performing unsupervised feature learning mapping the raw input image to a learned representation. Second, a \emph{Recurrent Neural Network (RNN)} performing language modeling on the textual description associated with the input picture. These approaches have the advantage of being differentiable end-to-end, thus allowing the use of gradient descent for optimization.

\begin{figure}[H]
    \begin{subfigure}{.5\textwidth}
        \centering
        \includegraphics[width=.94\linewidth]{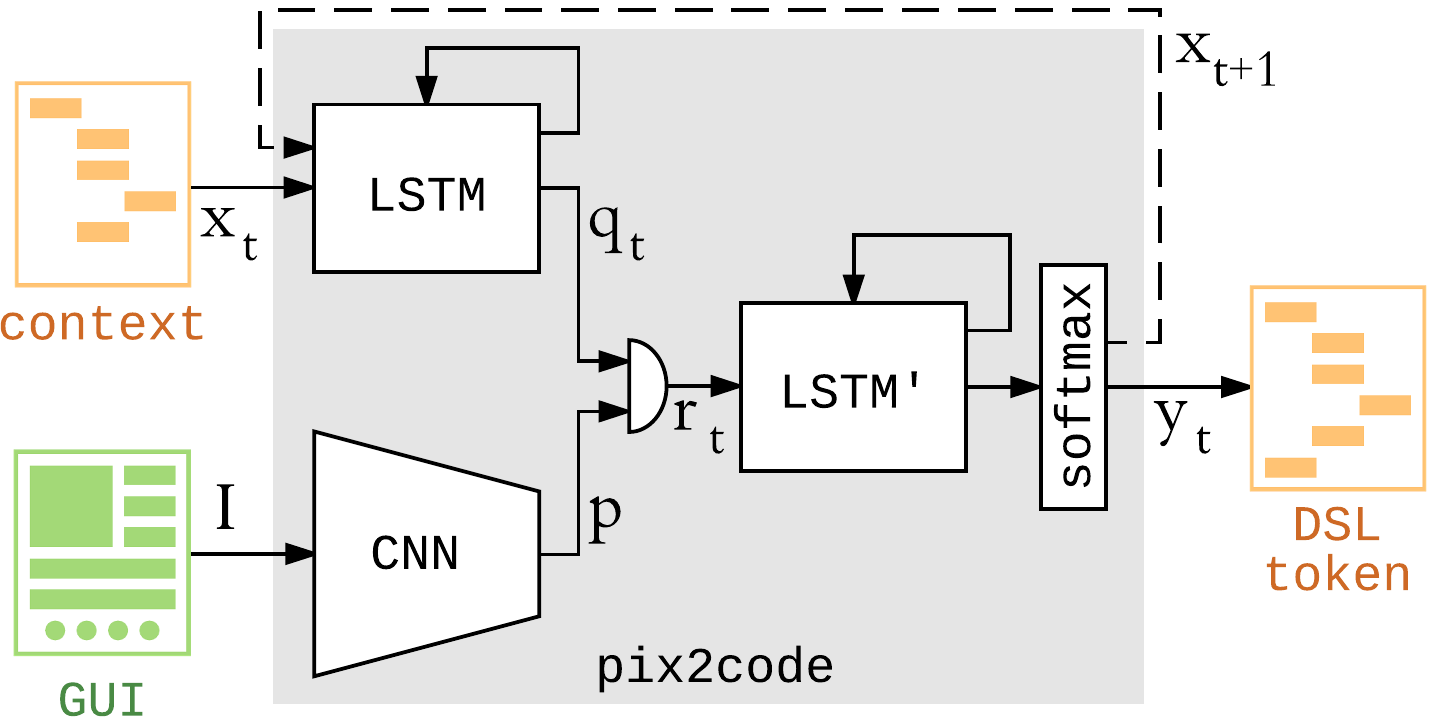}
        \caption{Training}
    \end{subfigure}
    \begin{subfigure}{.5\textwidth}
        \centering
        \includegraphics[width=.97\linewidth]{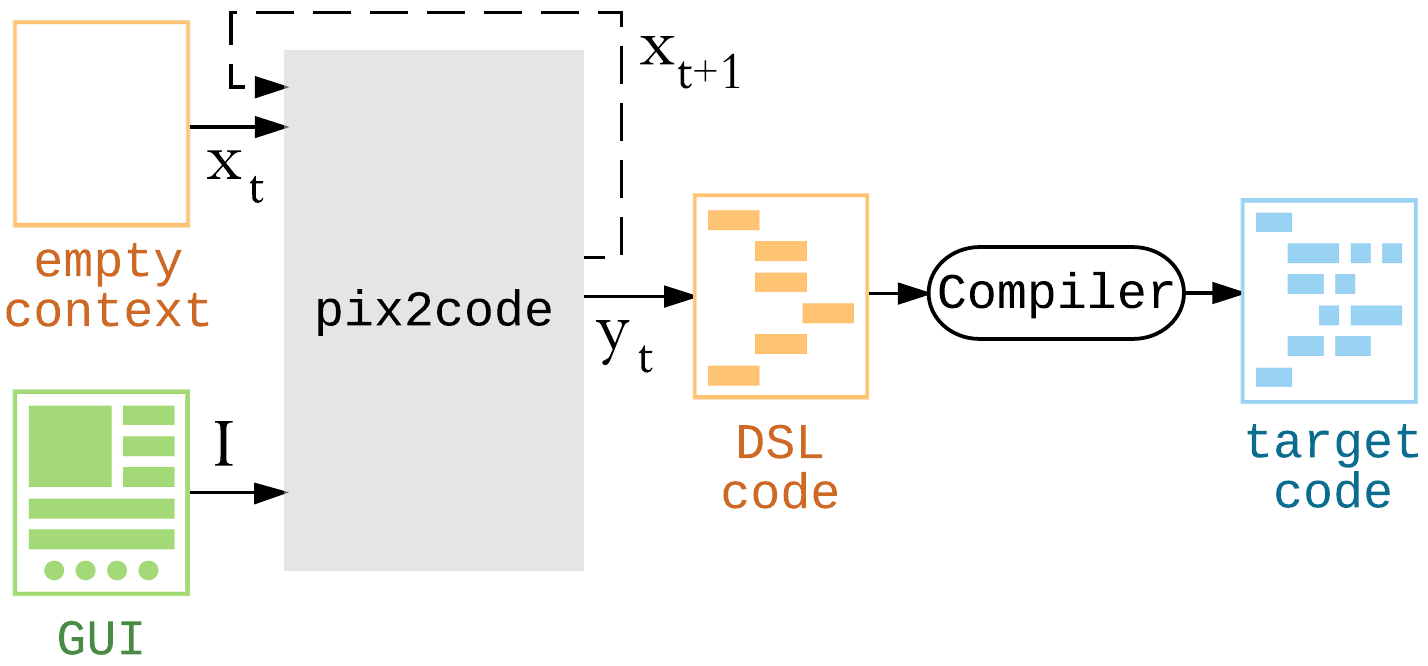}
        \caption{Sampling}
    \end{subfigure}
    \caption{Overview of the \emph{pix2code} model architecture. During training, the GUI image is encoded by a CNN-based vision model; the context (i.e. a sequence of one-hot encoded tokens corresponding to DSL code) is encoded by a language model consisting of a stack of LSTM layers. The two resulting feature vectors are then concatenated and fed into a second stack of LSTM layers acting as a decoder. Finally, a \emph{softmax} layer is used to sample one token at a time; the output size of the \emph{softmax} layer corresponding to the DSL vocabulary size. Given an image and a sequence of tokens, the model (i.e. contained in the gray box) is differentiable and can thus be optimized end-to-end through gradient descent to predict the next token in the sequence. During sampling, the input context is updated for each prediction to contain the last predicted token. The resulting sequence of DSL tokens is compiled to the desired target language using traditional compiler design techniques.}
    \label{fig:training_sampling}
\end{figure}

\section{pix2code}

The task of generating computer code written in a given programming language from a GUI screenshot can be compared to the task of generating English textual descriptions from a scene photography. In both scenarios, we want to produce a variable-length strings of tokens from pixel values. We can thus divide our problem into three sub-problems. First, a computer vision problem of understanding the given scene (i.e. in this case, the GUI image) and inferring the objects present, their identities, positions, and poses (i.e. buttons, labels, element containers). Second, a language modeling problem of understanding text (i.e. in this case, computer code) and generating syntactically and semantically correct samples. Finally, the last challenge is to use the solutions to both previous sub-problems by exploiting the latent variables inferred from scene understanding to generate corresponding textual descriptions (i.e. computer code rather than English) of the objects represented by these variables.

\subsection{Vision Model}

CNNs are currently the method of choice to solve a wide range of vision problems thanks to their topology allowing them to learn rich latent representations from the images they are trained on \cite{sermanet2013overfeat, krizhevsky2012imagenet}. We used a CNN to perform unsupervised feature learning by mapping an input image to a learned fixed-length vector; thus acting as an encoder as shown in Figure \ref{fig:training_sampling}.

The input images are initially re-sized to $256\times256$ pixels (the aspect ratio is not preserved) and the pixel values are normalized before to be fed in the CNN. No further pre-processing is performed. To encode each input image to a fixed-size output vector, we exclusively used small $3\times3$ receptive fields which are convolved with stride $1$ as used by Simonyan and Zisserman for VGGNet \cite{simonyan2014very}. These operations are applied twice before to down-sample with max-pooling. The width of the first convolutional layer is $32$, followed by a layer of width $64$, and finally width $128$. Two fully connected layers of size $1024$ applying the \emph{rectified linear unit} activation complete the vision model.

\begin{figure}[H]
    \begin{subfigure}{.5\textwidth}
        \centering
        \includegraphics[width=.41\linewidth]{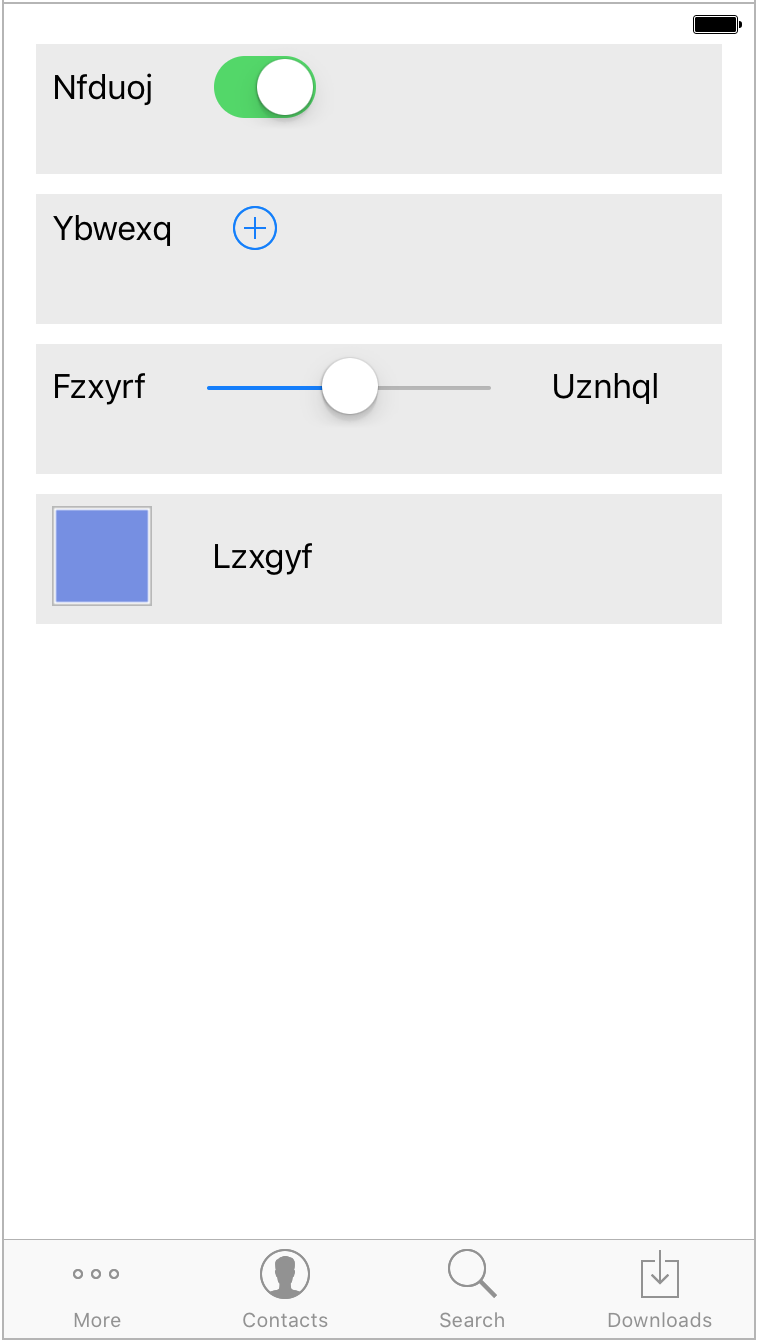}
        \caption{iOS GUI screenshot}
    \end{subfigure}
    \begin{subfigure}{.5\textwidth}
        \centering
        \includegraphics[width=.8\linewidth]{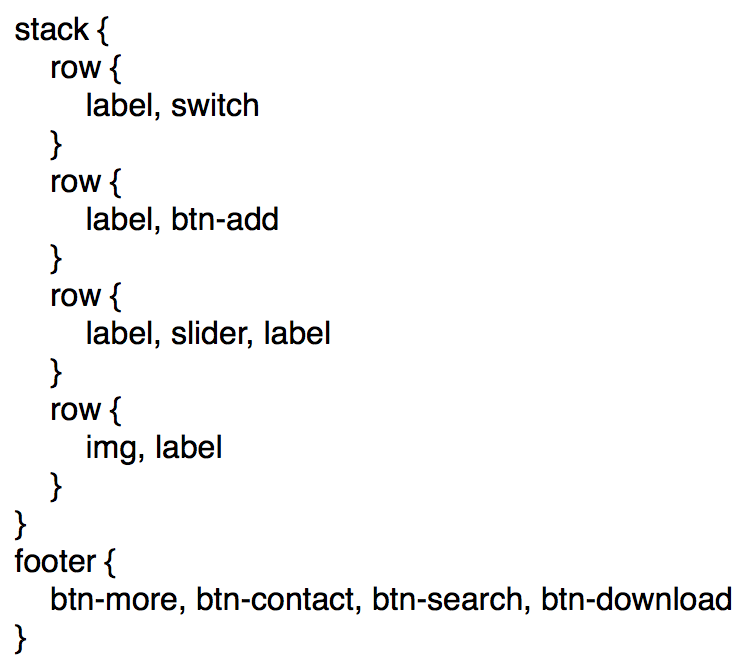}
        \caption{Code describing the GUI written in our DSL}
    \end{subfigure}
    \caption{An example of a native iOS GUI written in our markup-like DSL.}
    \label{fig:dsl}
\end{figure}

\subsection{Language Model}

We designed a simple lightweight DSL to describe GUIs as illustrated in Figure \ref{fig:dsl}. In this work we are only interested in the GUI layout, the different graphical components, and their relationships; thus the actual textual value of the labels is ignored. Additionally to reducing the size of the search space, the DSL simplicity also reduces the size of the vocabulary (i.e. the total number of tokens supported by the DSL). As a result, our language model can perform token-level language modeling with a discrete input by using one-hot encoded vectors; eliminating the need for word embedding techniques such as word2vec \cite{mikolov2013distributed} that can result in costly computations.

In most programming languages and markup languages, an element is declared with an opening token; if children elements or instructions are contained within a block, a closing token is usually needed for the interpreter or the compiler. In such a scenario where the number of children elements contained in a parent element is variable, it is important to model long-term dependencies to be able to close a block that has been opened.
Traditional RNN architectures suffer from vanishing and exploding gradients preventing them from being able to model such relationships between data points spread out in time series (i.e. in this case tokens spread out in a sequence). Hochreiter and Schmidhuber proposed the \emph{Long Short-Term Memory (LSTM)} neural architecture in order to address this very problem \cite{hochreiter1997long}. The different LSTM gate outputs can be computed as follows:

\begin{align}
    i_t &= \phi(W_{ix}x_{t} + W_{iy}h_{t-1} + b_i)\\
    f_t &= \phi(W_{fx}x_{t} + W_{fy}h_{t-1} + b_f)\\
    o_t &= \phi(W_{ox}x_{t} + W_{oy}h_{t-1} + b_o)\\
    c_t &= f_{t}\bullet c_{t-1} + i_{t}\bullet \sigma(W_{cx}x_{t} + W_{cy}h_{t-1} + b_c)\\
    h_t &= o_{t}\bullet \sigma(c_{t})
\end{align}

With $W$ the matrices of weights, $x_t$ the new input vector at time $t$, $h_{t-1}$ the previously produced output vector, $c_{t-1}$ the previously produced cell state's output, $b$ the biases, and $\phi$ and $\sigma$ the activation functions \emph{sigmoid} and \emph{hyperbolic tangent}, respectively. The cell state $c$ learns to memorize information by using a recursive connection as done in traditional RNN cells. The input gate $i$ is used to control the error flow on the inputs of cell state $c$ to avoid input weight conflicts that occur in traditional RNN because the same weight has to be used for both storing certain inputs and ignoring others. The output gate $o$ controls the error flow from the outputs of the cell state $c$ to prevent output weight conflicts that happen in standard RNN because the same weight has to be used for both retrieving information and not retrieving others. The LSTM memory block can thus use $i$ to decide when to write information in $c$ and use $o$ to decide when to read information from $c$. We used the LSTM variant proposed by Gers and Schmidhuber \cite{gers2000learning} with a forget gate $f$ to reset memory and help the network model continuous sequences.

\subsection{Decoder}

Our model is trained in a supervised learning manner by feeding an image $I$ and a contextual sequence $X$ of $T$ tokens $x_t, t\in \left\{ 0\ldots T-1 \right\}$ as inputs; and the token $x_{T}$ as the target label.
As shown on Figure \ref{fig:training_sampling}, a CNN-based vision model encodes the input image $I$ into a vectorial representation $p$. The input token $x_t$ is encoded by an LSTM-based language model into an intermediary representation $q_t$ allowing the model to focus more on certain tokens and less on others \cite{graves2013generating}. This first language model is implemented as a stack of two LSTM layers with $128$ cells each.
The vision-encoded vector $p$ and the language-encoded vector $q_t$ are concatenated into a single feature vector $r_t$ which is then fed into a second LSTM-based model decoding the representations learned by both the vision model and the language model. The decoder thus learns to model the relationship between objects present in the input GUI image and the associated tokens present in the DSL code. Our decoder is implemented as a stack of two LSTM layers with $512$ cells each. This architecture can be expressed mathematically as follows:

\begin{align}
p &= CNN(I)\\
q_t &= LSTM(x_t)\\
r_t &= (q, p_t)\\
y_t &= softmax(LSTM'(r_t))\\
x_{t+1} &= y_t 
\end{align}

This architecture allows the whole \emph{pix2code} model to be optimized end-to-end with gradient descent to predict a token at a time after it has seen both the image as well as the preceding tokens in the sequence.
The discrete nature of the output (i.e. fixed-sized vocabulary of tokens in the DSL) allows us to reduce the task to a classification problem. That is, the output layer of our model has the same number of cells as the vocabulary size; thus generating a probability distribution of the candidate tokens at each time step allowing the use of a \emph{softmax} layer to perform multi-class classification.

\subsection{Training}

The length $T$ of the sequences used for training is important to model long-term dependencies; for example to be able to close a block of code that has been opened. After conducting empirical experiments, the DSL input files used for training were segmented with a sliding window of size $48$; in other words, we unroll the recurrent neural network for $48$ steps. This was found to be a satisfactory trade-off between long-term dependencies learning and computational cost. For every token in the input DSL file, the model is therefore fed with both an input image and a contextual sequence of $T=48$ tokens. While the context (i.e. sequence of tokens) used for training is updated at each time step (i.e. each token) by sliding the window, the very same input image $I$ is reused for samples associated with the same GUI. The special tokens $<START>$ and $<END>$ are used to respectively prefix and suffix the DSL files similarly to the method used by Karpathy and Fei-Fei \cite{karpathy2015deep}. Training is performed by computing the partial derivatives of the loss with respect to the network weights calculated with backpropagation to minimize the multiclass log loss:

\begin{align}
L(I, X) &= - \sum^{T}_{t=1} x_{t+1} \log (y_{t})
\end{align}

With $x_{t+1}$ the expected token, and $y_{t}$ the predicted token. The model is optimized end-to-end hence the loss $L$ is minimized with regard to all the parameters including all layers in the CNN-based vision model and all layers in both LSTM-based models. Training with the \emph{RMSProp} algorithm \cite{tieleman2012lecture} gave the best results with a learning rate set to $1e-4$ and by clipping the output gradient to the range $[-1.0, 1.0]$ to cope with numerical instability \cite{graves2013generating}. To prevent overfitting, a dropout regularization \cite{srivastava2014dropout} set to $25\%$ is applied to the vision model after each max-pooling operation and at $30\%$ after each fully-connected layer. In the LSTM-based models, dropout is set to $10\%$ and only applied to the non-recurrent connections \cite{zaremba2014recurrent}. Our model was trained with mini-batches of $64$ image-sequence pairs.

\begin{table}[H]
    \caption{Dataset statistics.}
    \centering
    \begin{tabular}{| c | c | c | c | c | c |} \hline
        \multirow{2}{*}{\textbf{Dataset type}} &
        \multirow{2}{*}{\textbf{Synthesizable}} &
        \multicolumn{2}{c|}{\textbf{Training set}} & \multicolumn{2}{c|}{\textbf{Test set}} \\ \cline{3-6}
        & & Instances & Samples & Instances & Samples \\ \hline
        iOS UI (Storyboard) & $26 \times 10^5$ & $1500$ & $93672$ & $250$ & $15984$ \\ \hline
        Android UI (XML) & $21 \times 10^6$ & $1500$ & $85756$ & $250$ & $14265$ \\ \hline
        web-based UI (HTML/CSS) & $31 \times 10^4$ & $1500$ & $143850$ & $250$ & $24108$ \\ \hline
    \end{tabular}
    \label{tab:dataset}
\end{table}

\subsection{Sampling}

To generate DSL code, we feed the GUI image $I$ and a contextual sequence $X$ of $T=48$ tokens where tokens $x_t \ldots x_{T-1}$ are initially set empty and the last token of the sequence $x_T$ is set to the special $<START>$ token. The predicted token $y_t$ is then used to update the next sequence of contextual tokens. That is, $x_t \ldots x_{T-1}$ are set to $x_{t+1} \ldots x_{T}$ ($x_t$ is thus discarded), with $x_T$ set to $y_t$. The process is repeated until the token $<END>$ is generated by the model.
The generated DSL token sequence can then be compiled with traditional compilation methods to the desired target language.

\begin{figure}[H]
    \begin{subfigure}{.5\textwidth}
        \centering
        \includegraphics[width=1.\linewidth]{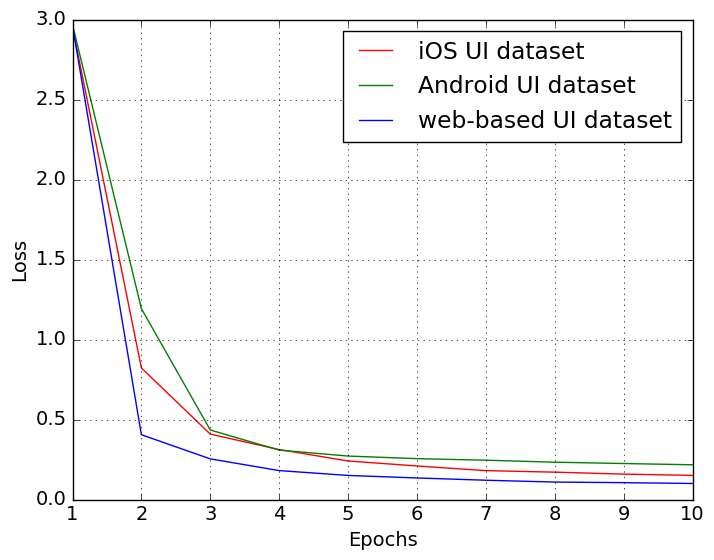}
        \caption{\emph{pix2code} training loss}
    \end{subfigure}
    \begin{subfigure}{.5\textwidth}
        \centering
        \includegraphics[width=1.\linewidth]{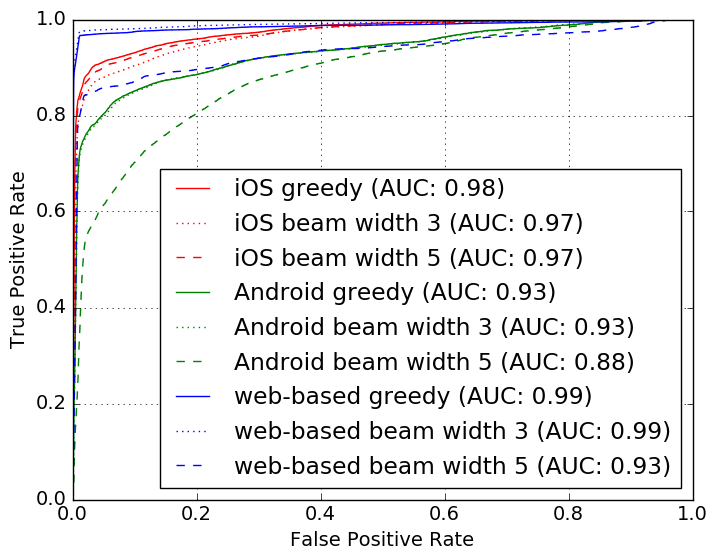}
        \caption{Micro-average ROC curves}
    \end{subfigure}
    \caption{Training loss on different datasets and ROC curves calculated during sampling with the model trained for 10 epochs.}
    \label{fig:training_stats}
\end{figure}

\section{Experiments}

Access to consequent datasets is a typical bottleneck when training deep neural networks. To the best of our knowledge, no dataset consisting of both GUI screenshots and source code was available at the time this paper was written. As a consequence, we synthesized our own data resulting in the three datasets described in Table \ref{tab:dataset}. The column \emph{Synthesizable} refers to the maximum number of unique GUI configuration that can be synthesized using our stochastic user interface generator. The columns \emph{Instances} refers to the number of synthesized (GUI screenshot, GUI code) file pairs. The columns \emph{Samples} refers to the number of distinct image-sequence pairs. In fact, training and sampling are done one token at a time by feeding the model with an image and a sequence of tokens obtained with a sliding window of fixed size $T$. The total number of training samples thus depends on the total number of tokens written in the DSL files and the size of the sliding window. Our stochastic user interface generator is designed to synthesize GUIs written in our DSL which is then compiled to the desired target language to be rendered. Using data synthesis also allows us to demonstrate the capability of our model to generate computer code for three different platforms.

\begin{table}[H]
    \caption{Experiments results reported for the test sets described in Table \ref{tab:dataset}.}
    \centering
    \begin{tabular}{| c | c | c | c |} \hline
        \multirow{2}{*}{\textbf{Dataset type}} &
        \multicolumn{3}{c|}{\textbf{Error (\%)}} \\ \cline{2-4}
        & \textbf{greedy search} & \textbf{beam search 3} & \textbf{beam search 5}\\ \hline
        iOS UI (Storyboard) & \textbf{22.73} & 25.22 & 23.94 \\ \hline
        Android UI (XML) & \textbf{22.34} & 23.58 & 40.93 \\ \hline
        web-based UI (HTML/CSS) & 12.14 & \textbf{11.01} & 22.35 \\ \hline
    \end{tabular}
    \label{tab:experiments}
\end{table}

Our model has around $109 \times 10^6$ parameters to optimize and all experiments are performed with the same model with no specific tuning; only the training datasets differ as shown on Figure \ref{fig:training_stats}. Code generation is performed with both greedy search and beam search to find the tokens that maximize the classification probability. To evaluate the quality of the generated output, the classification error is computed for each sampled DSL token and averaged over the whole test dataset. The length difference between the generated and the expected token sequences is also counted as error. The results can be seen on Table \ref{tab:experiments}.

\begin{figure}[H]
    \begin{subfigure}{.245\textwidth}
        \centering
        \includegraphics[width=.9\linewidth]{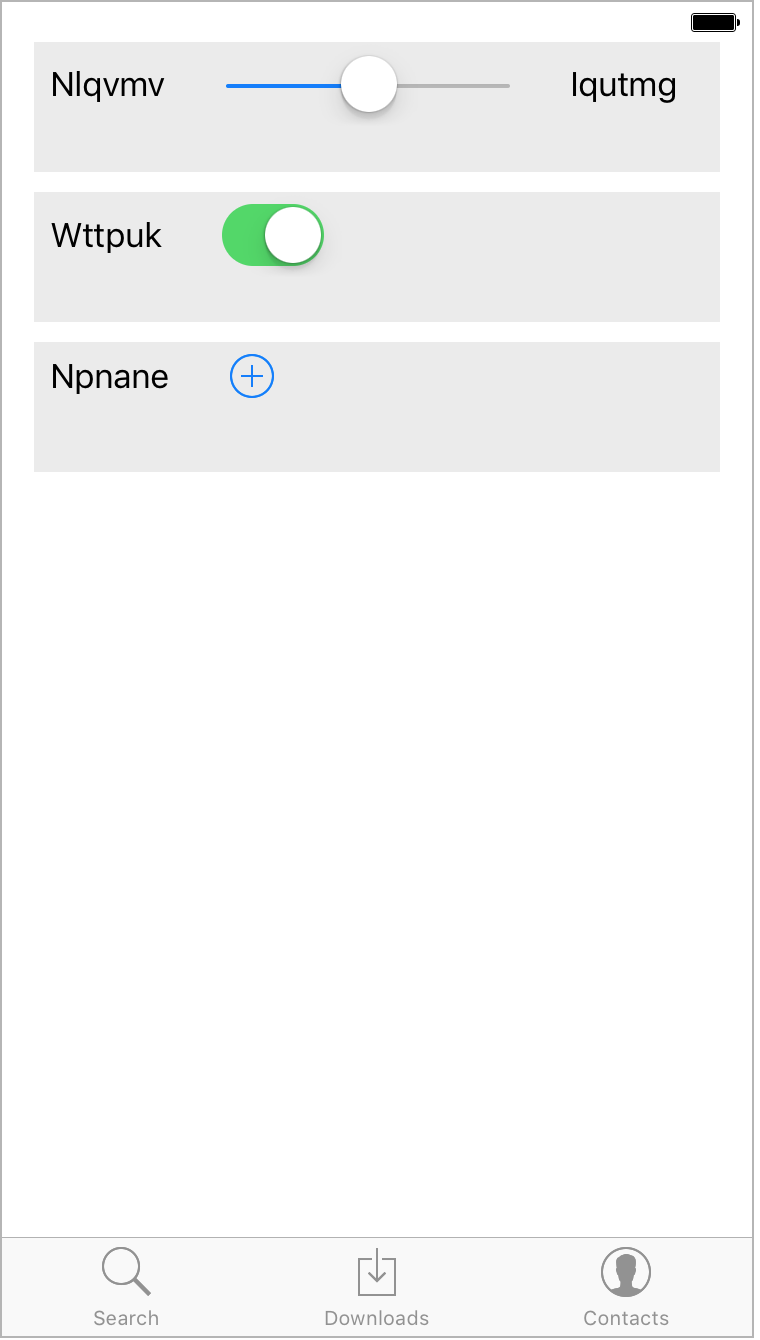}
        \caption{Groundtruth GUI 1}
    \end{subfigure}
    \begin{subfigure}{.245\textwidth}
        \centering
        \includegraphics[width=.9\linewidth]{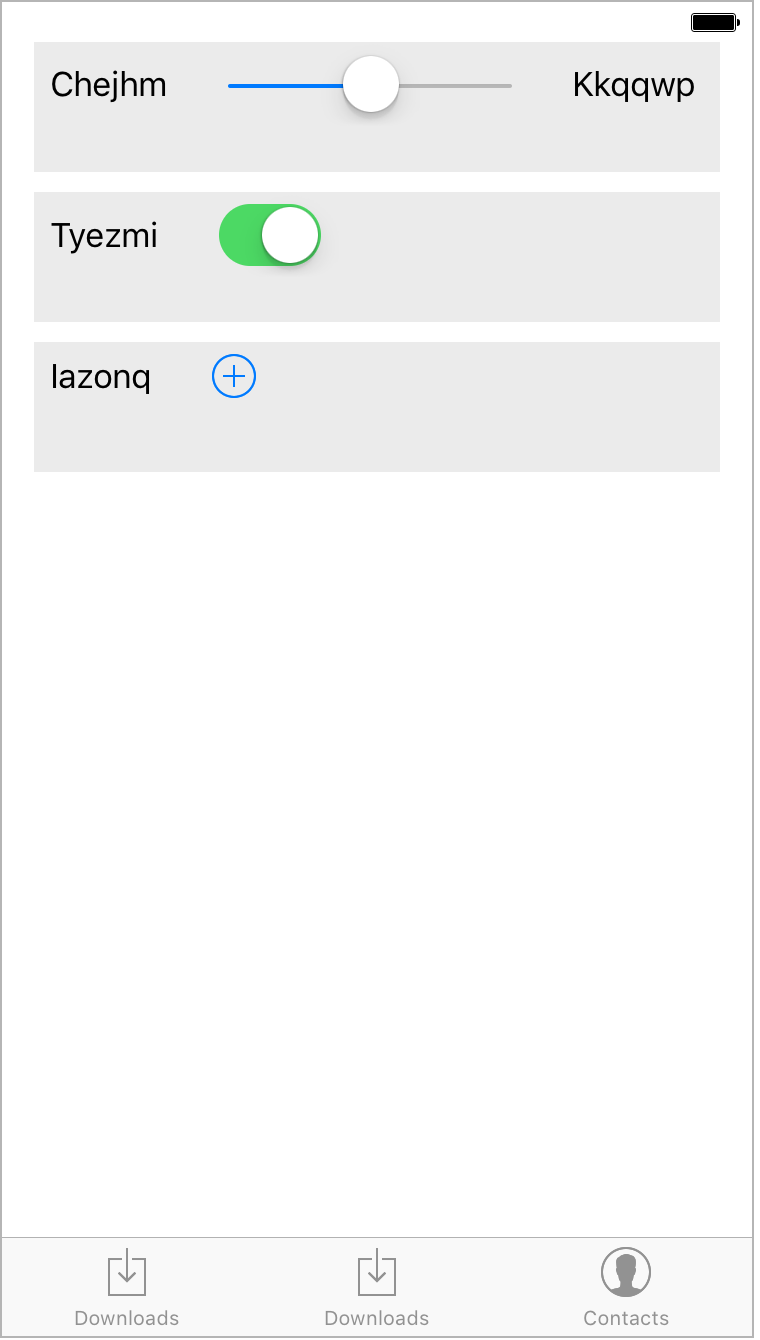}
        \caption{Generated GUI 1}
    \end{subfigure}
    \begin{subfigure}{.245\textwidth}
        \centering
        \includegraphics[width=.9\linewidth]{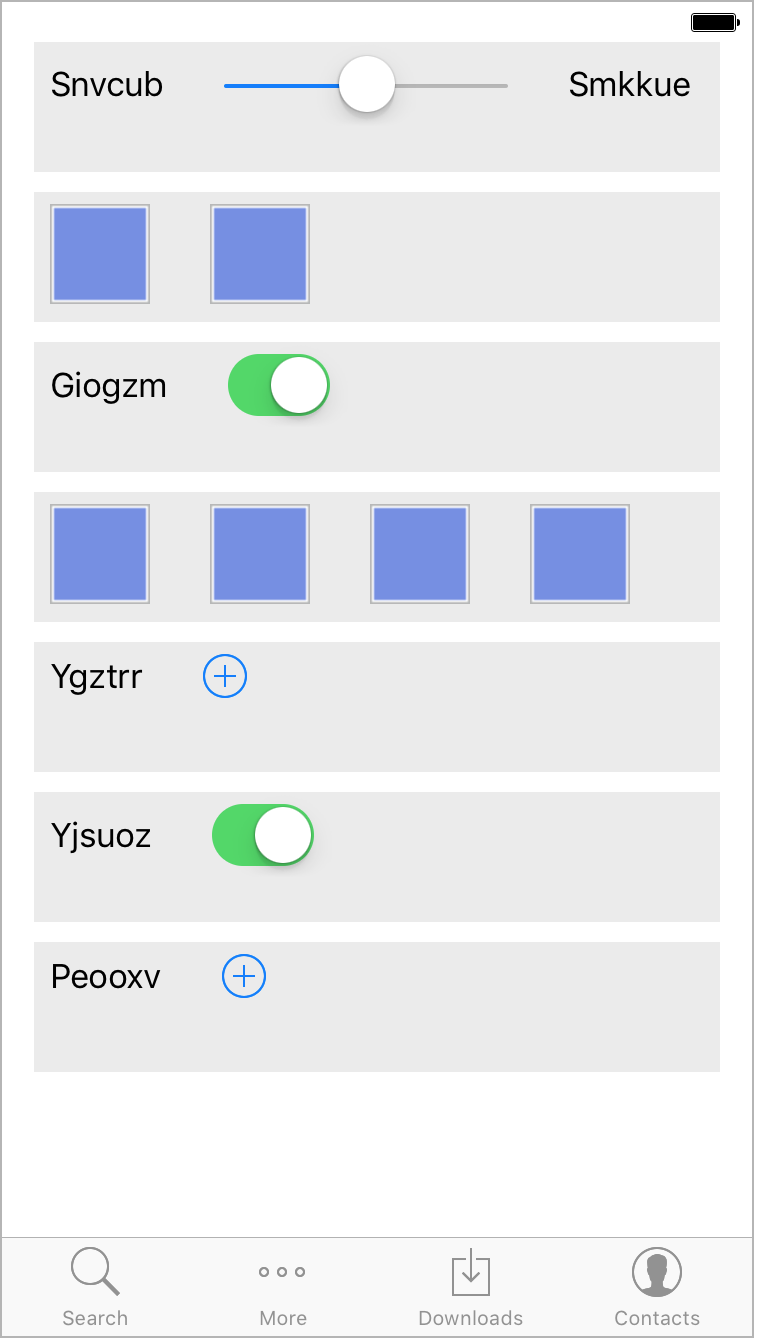}
        \caption{Groundtruth GUI 2}
    \end{subfigure}
    \begin{subfigure}{.245\textwidth}
        \centering
        \includegraphics[width=.9\linewidth]{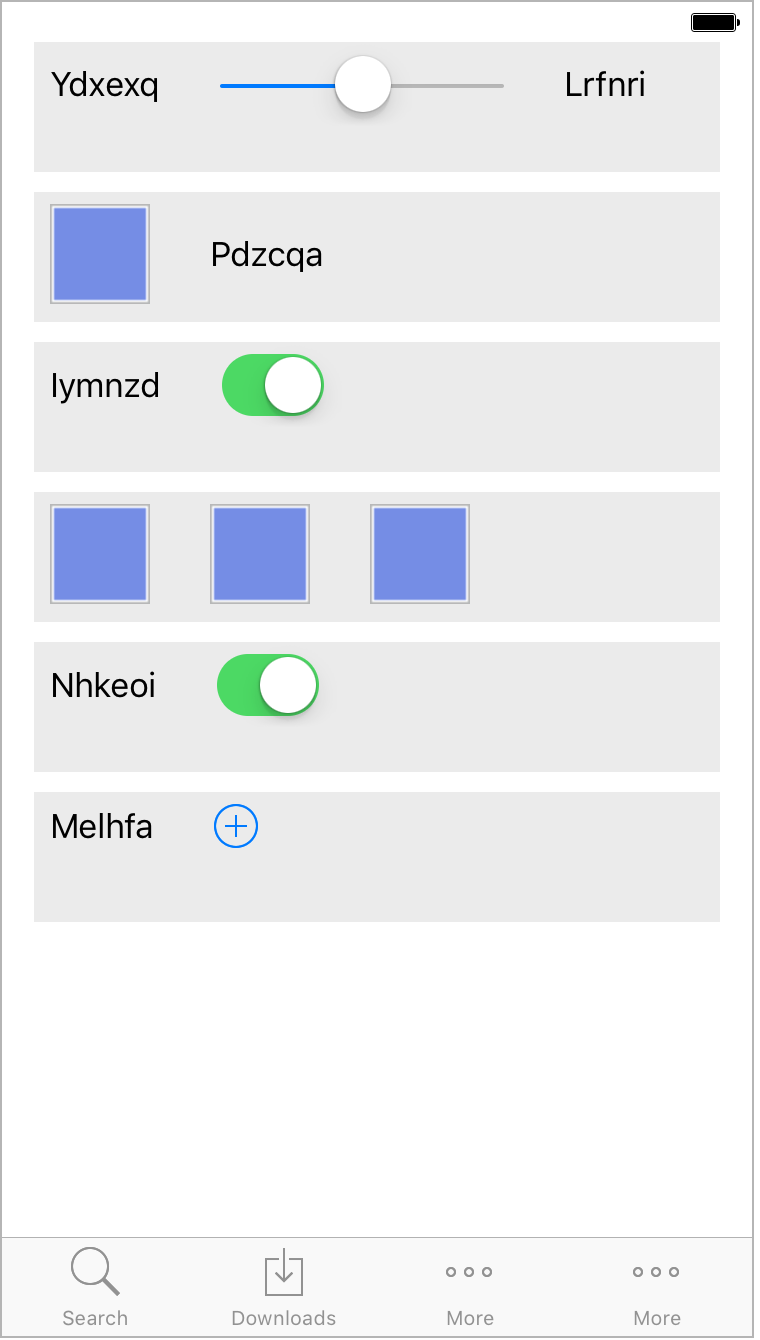}
        \caption{Generated GUI 2}
    \end{subfigure}
    \caption{Experiment samples for the iOS GUI dataset.}
    \label{fig:samples_ios}
\end{figure}

\begin{figure}[H]
    \begin{subfigure}{.495\textwidth}
        \centering
        \includegraphics[width=1.\linewidth]{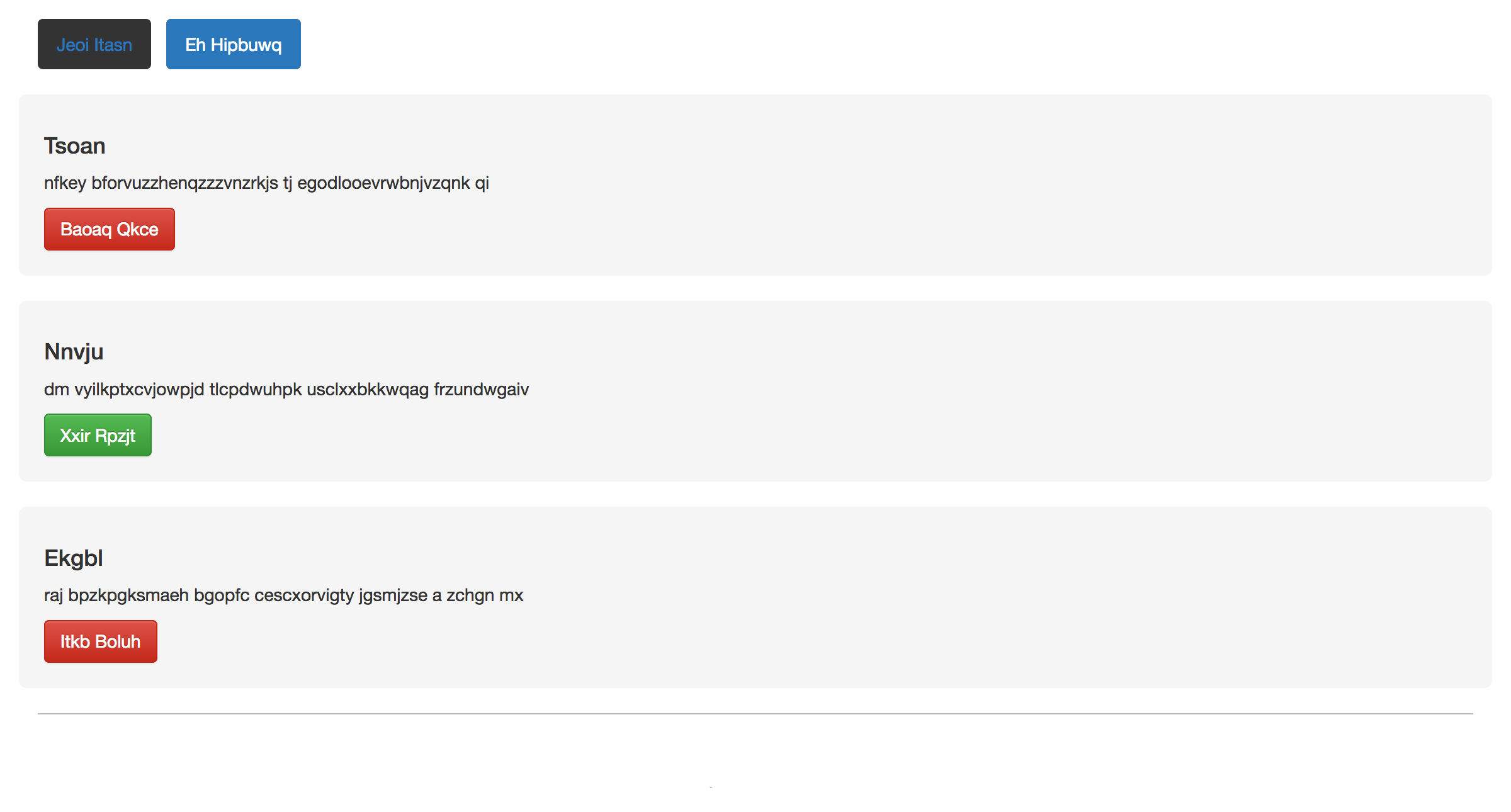}
        \caption{Groundtruth GUI 5}
    \end{subfigure}
    \begin{subfigure}{.495\textwidth}
        \centering
        \includegraphics[width=1.\linewidth]{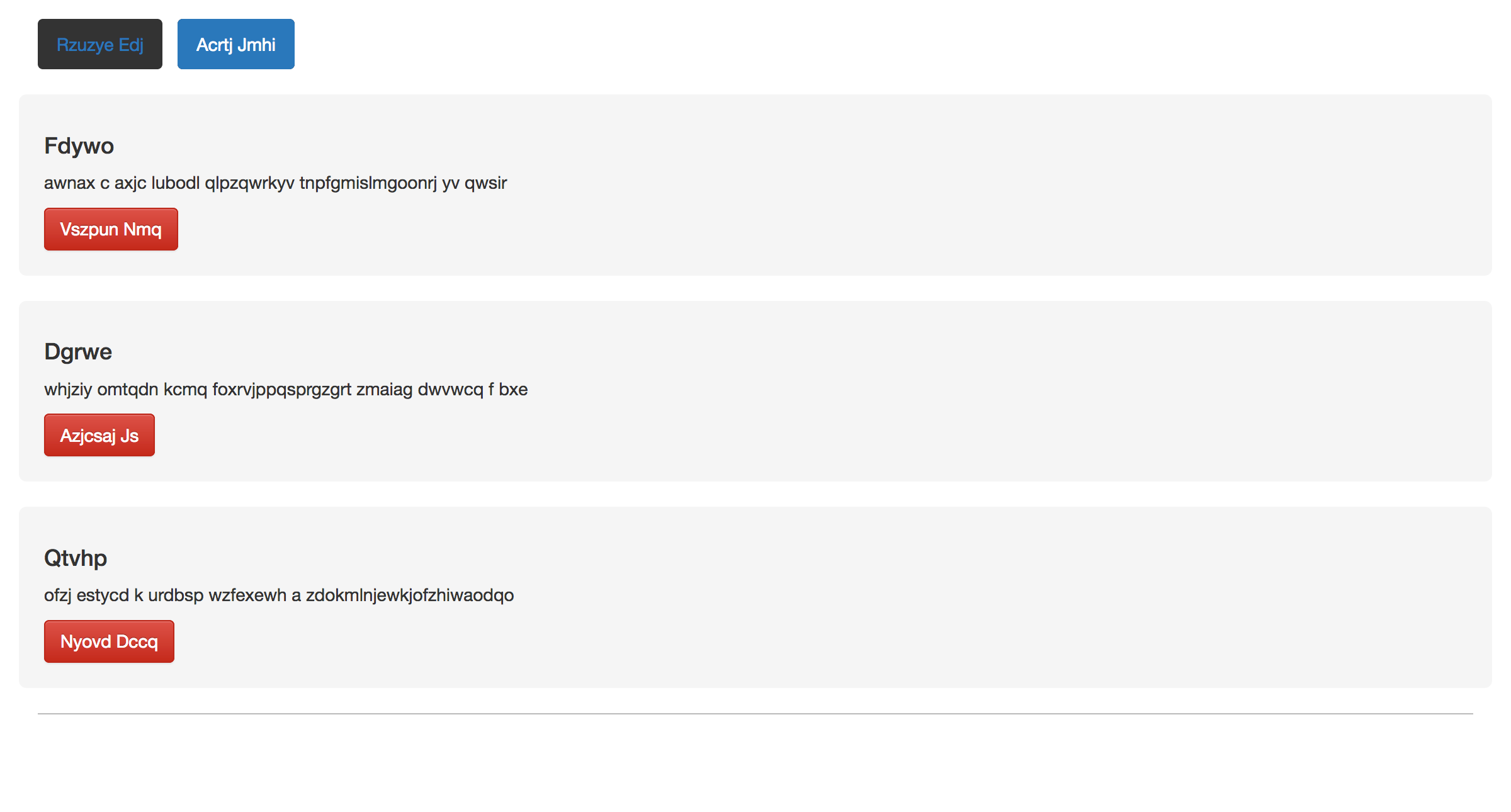}
        \caption{Generated GUI 5}
    \end{subfigure}
    \begin{subfigure}{.495\textwidth}
        \centering
        \includegraphics[width=1.\linewidth]{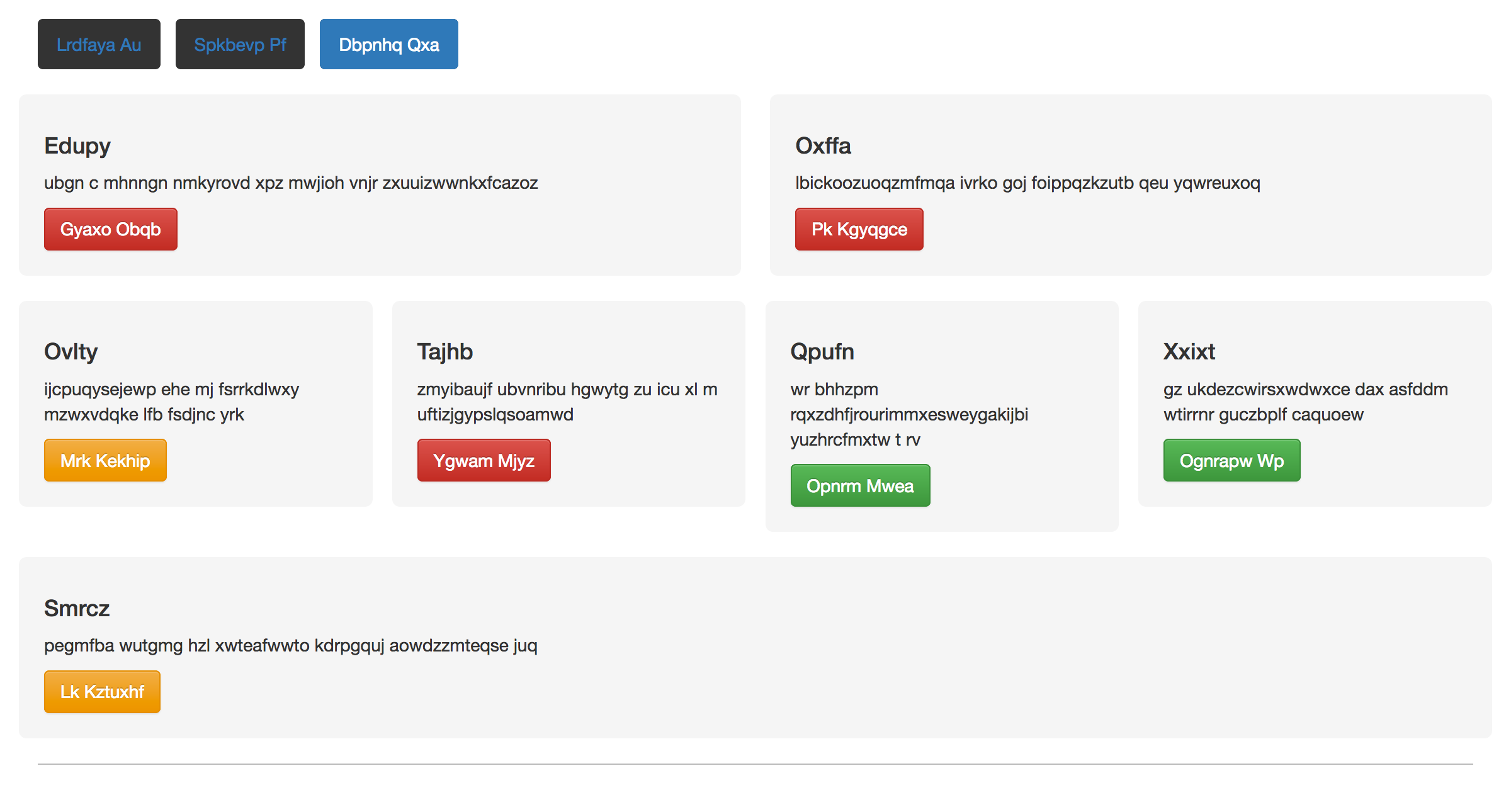}
        \caption{Groundtruth GUI 6}
    \end{subfigure}
    \begin{subfigure}{.495\textwidth}
        \centering
        \includegraphics[width=1.\linewidth]{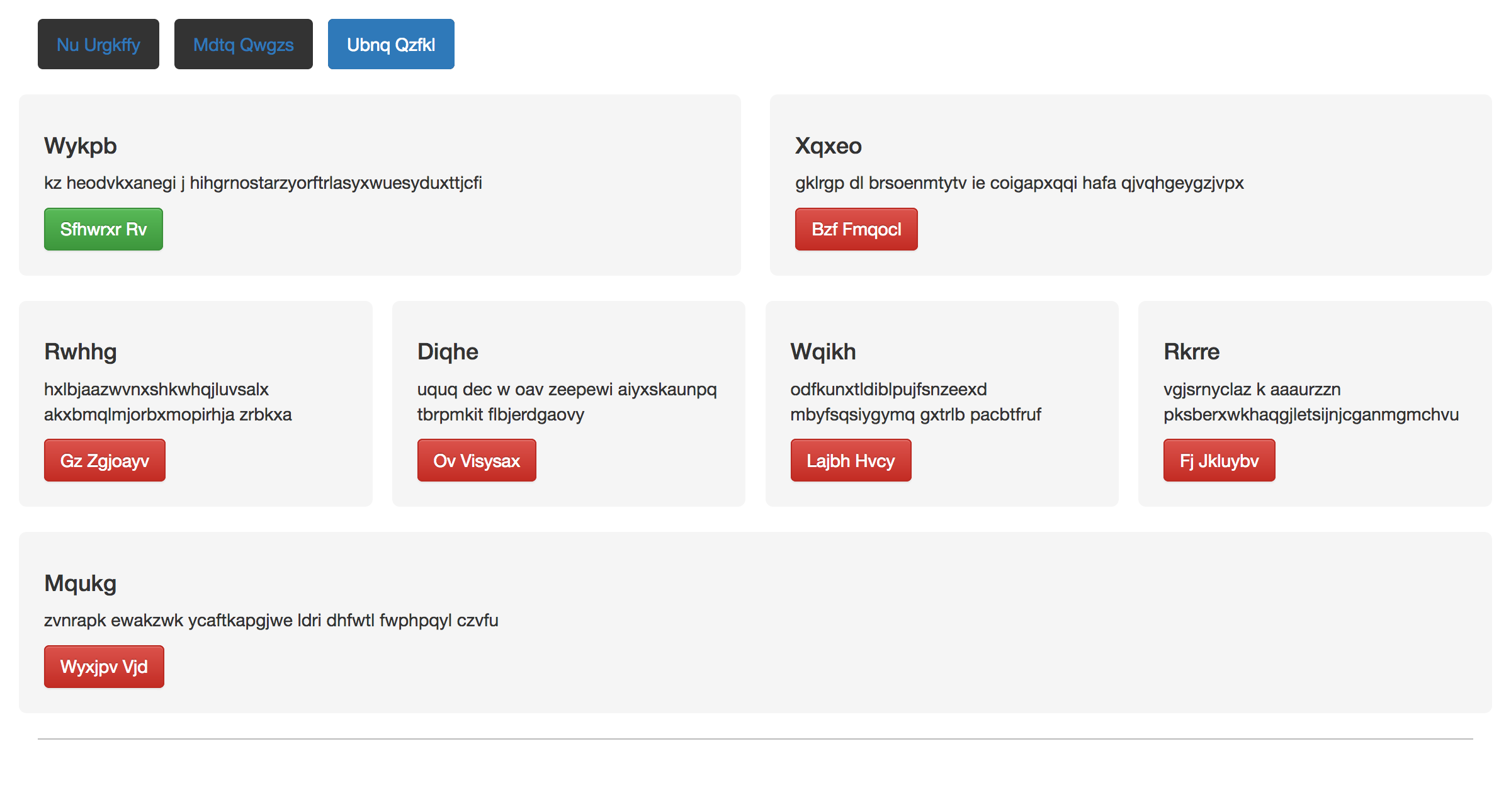}
        \caption{Generated GUI 6}
    \end{subfigure}
    \caption{Experiment samples from the web-based GUI dataset.}
    \label{fig:samples_web}
\end{figure}

\begin{figure}[H]
    \begin{subfigure}{.245\textwidth}
        \centering
        \includegraphics[width=.9\linewidth]{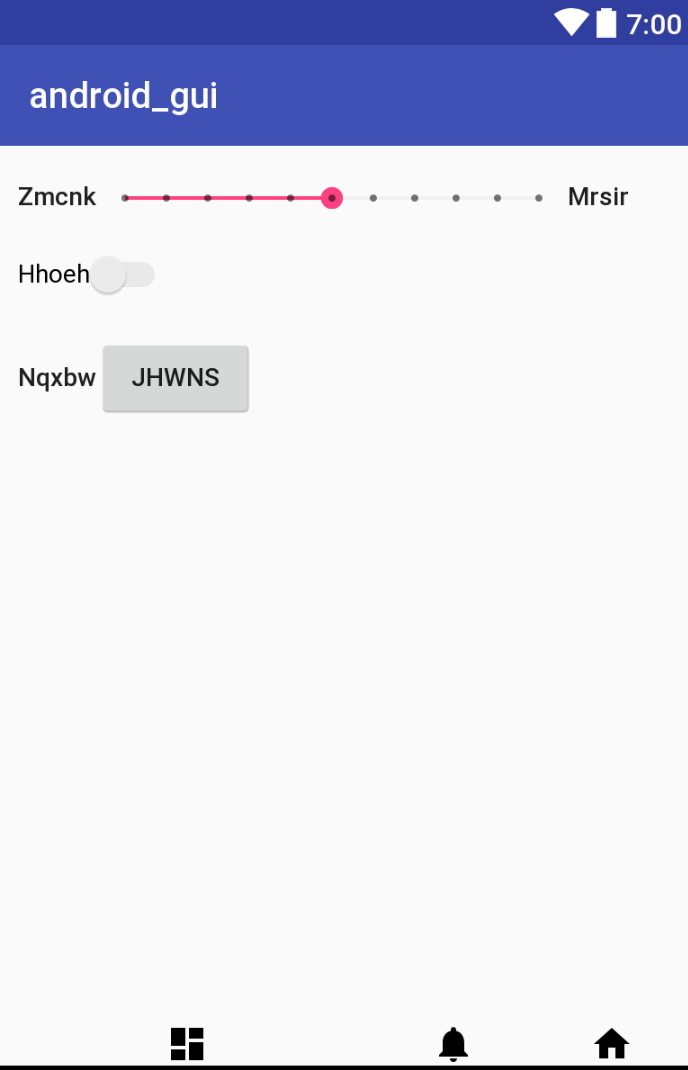}
        \caption{Groundtruth GUI 3}
    \end{subfigure}
    \begin{subfigure}{.245\textwidth}
        \centering
        \includegraphics[width=.9\linewidth]{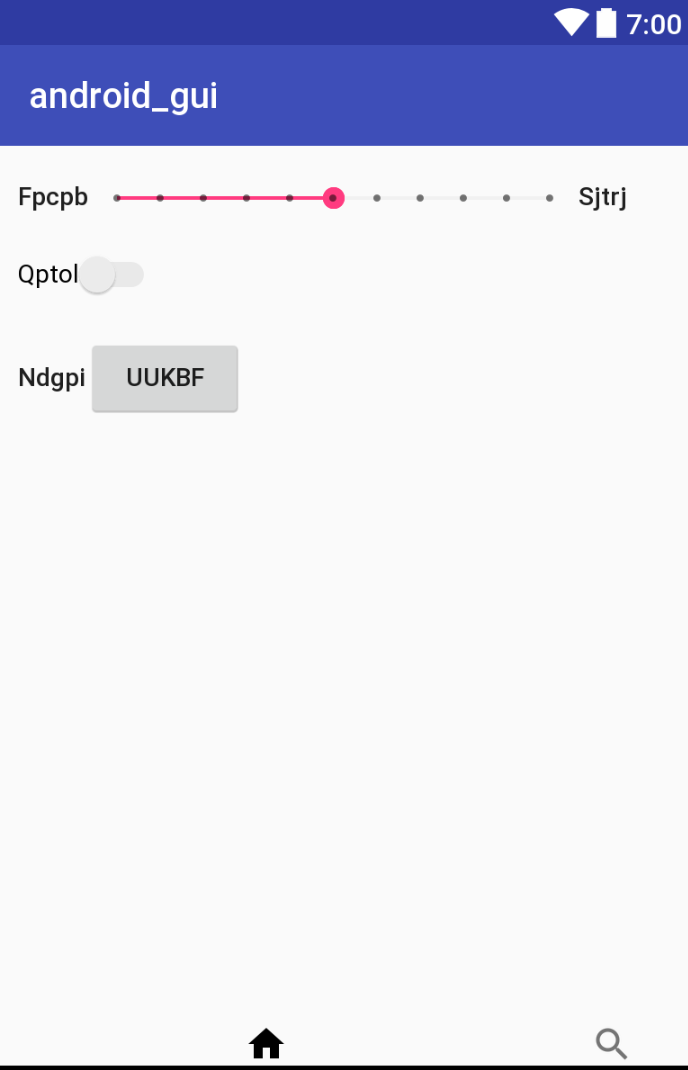}
        \caption{Generated GUI 3}
    \end{subfigure}
    \begin{subfigure}{.245\textwidth}
        \centering
        \includegraphics[width=.9\linewidth]{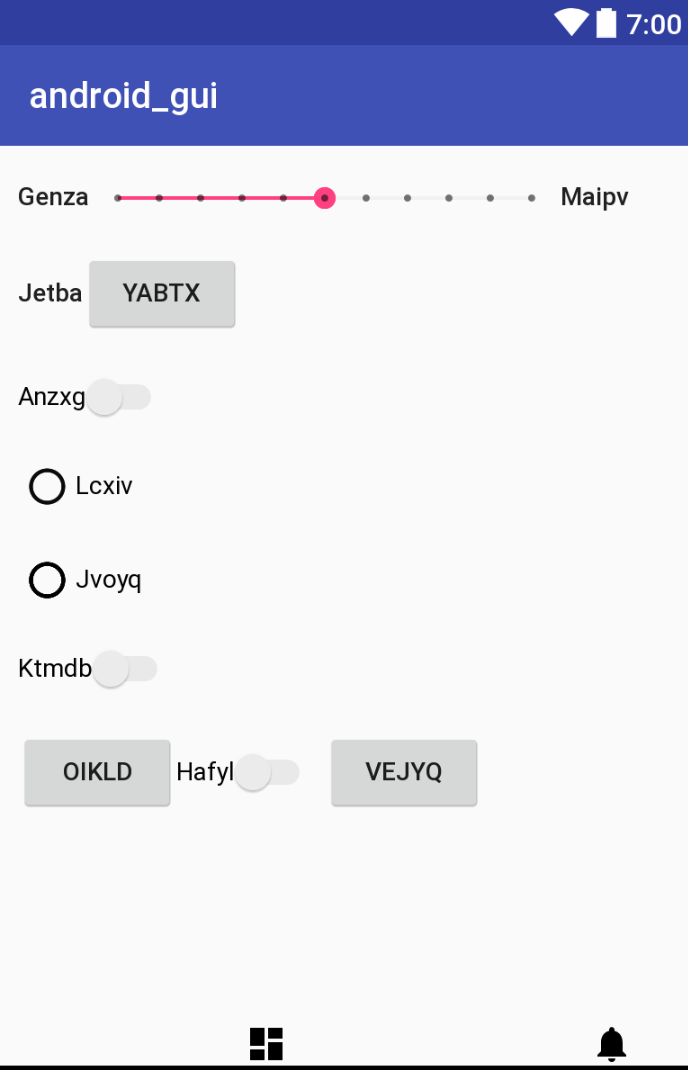}
        \caption{Groundtruth GUI 4}
    \end{subfigure}
    \begin{subfigure}{.245\textwidth}
        \centering
        \includegraphics[width=.9\linewidth]{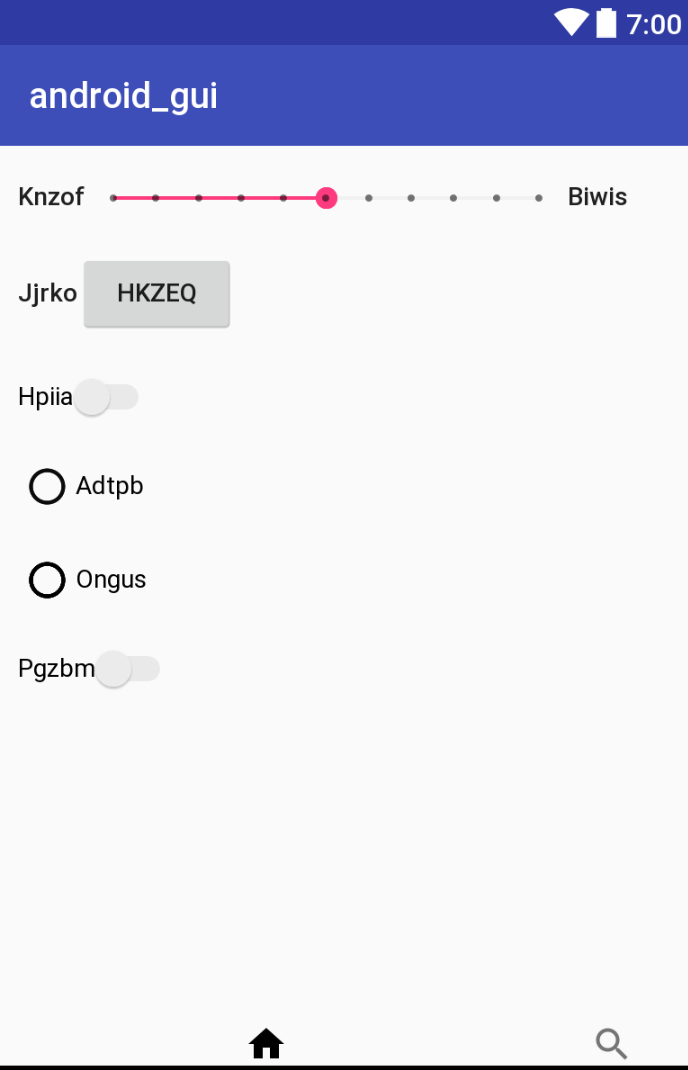}
        \caption{Generated GUI 4}
    \end{subfigure}
    \caption{Experiment samples from the Android GUI dataset.}
    \label{fig:samples_android}
\end{figure}

Figures \ref{fig:samples_ios}, \ref{fig:samples_web}, and \ref{fig:samples_android} show samples consisting of input GUIs (i.e. ground truth), and output GUIs generated by a trained \emph{pix2code} model. It is important to remember that the actual textual value of the labels is ignored and that both our data synthesis algorithm and our DSL compiler assign randomly generated text to the labels. Despite occasional problems to select the right color or the right style for specific GUI elements and some difficulties modelling GUIs consisting of long lists of graphical components, our model is generally able to learn the GUI layout in a satisfying manner and can preserve the hierarchical structure of the graphical elements.


\section{Conclusion and Discussions}

In this paper, we presented \emph{pix2code}, a novel method to generate computer code given a single GUI image as input. While our work demonstrates the potential of such a system to automate the process of implementing GUIs, we only scratched the surface of what is possible. Our model consists of relatively few parameters and was trained on a relatively small dataset. The quality of the generated code could be drastically improved by training a bigger model on significantly more data for an extended number of epochs. Implementing a now-standard attention mechanism \cite{bahdanau2014neural, xu2015show} could further improve the quality of the generated code.

Using one-hot encoding does not provide any useful information about the relationships between the tokens since the method simply assigns an arbitrary vectorial representation to each token. Therefore, pre-training the language model to learn vectorial representations would allow the relationships between tokens in the DSL to be inferred (i.e. learning word embeddings such as word2vec \cite{mikolov2013distributed}) and as a result alleviate semantical error in the generated code. Furthermore, one-hot encoding does not scale to very big vocabulary and thus restrict the number of symbols that the DSL can support. 

\emph{Generative Adversarial Networks GANs} \cite{goodfellow2014generative} have shown to be extremely powerful at generating images and sequences \cite{yu2016seqgan, reed2016generative, zhang2016stackgan, shetty2017speaking, dai2017towards}. Applying such techniques to the problem of generating computer code from an input image is so far an unexplored research area. GANs could potentially be used as a standalone method to generate code or could be used in combination with our \emph{pix2code} model to fine-tune results.

A major drawback of deep neural networks is the need for a lot of training data for the resulting model to generalize well on new unseen examples. One of the significant advantages of the method we described in this paper is that there is no need for human-labelled data. In fact, the network can model the relationships between graphical components and associated tokens by simply being trained on image-sequence pairs. Although we used data synthesis in our paper partly to demonstrate the capability of our method to generate GUI code for various platforms; data synthesis might not be needed at all if one wants to focus only on web-based GUIs. In fact, one could imagine crawling the World Wide Web to collect a dataset of HTML/CSS code associated with screenshots of rendered websites. Considering the large number of web pages already available online and the fact that new websites are created every day, the web could theoretically supply a virtually unlimited amount of training data; potentially allowing deep learning methods to fully automate the implementation of web-based GUIs.

\bibliographystyle{abbrv}
\bibliography{references}

\end{document}